\documentclass{article}
\usepackage{spconf,amsmath,graphicx}
\usepackage{amssymb}
\usepackage{booktabs}
\graphicspath{{figs/}}
\usepackage{color, colortbl}

\usepackage{graphicx}
\usepackage{amsmath}
\usepackage{amssymb}
\usepackage{amsthm}
\usepackage{booktabs}
\usepackage{mathtools}
\usepackage{algorithm}
\usepackage{algpseudocode}

\usepackage{graphicx}
\usepackage{caption}

\title{On-the-fly object detection using StyleGAN with CLIP guidance}
\name{Yuzhe Lu$^{\dagger}\sthanks{Work done 
 	while at Lawrence Livermore National Laboratory}$, Shusen Liu$^{\star}$, Jayaraman J. Thiagarajan$^{\star}$, Wesam Sakla$^{\star}$, Rushil Anirudh$^{\star}$}

\address{$^{\dagger}$MLD, Carnegie Mellon University \\ $^{\star}$Lawrence Livermore National Laboratory}

\let\oldtwocolumn\twocolumn
\renewcommand\twocolumn[1][]{%
    \oldtwocolumn[{#1}{
    \begin{center}
        \vspace{-5mm}
        \includegraphics[width=\textwidth]{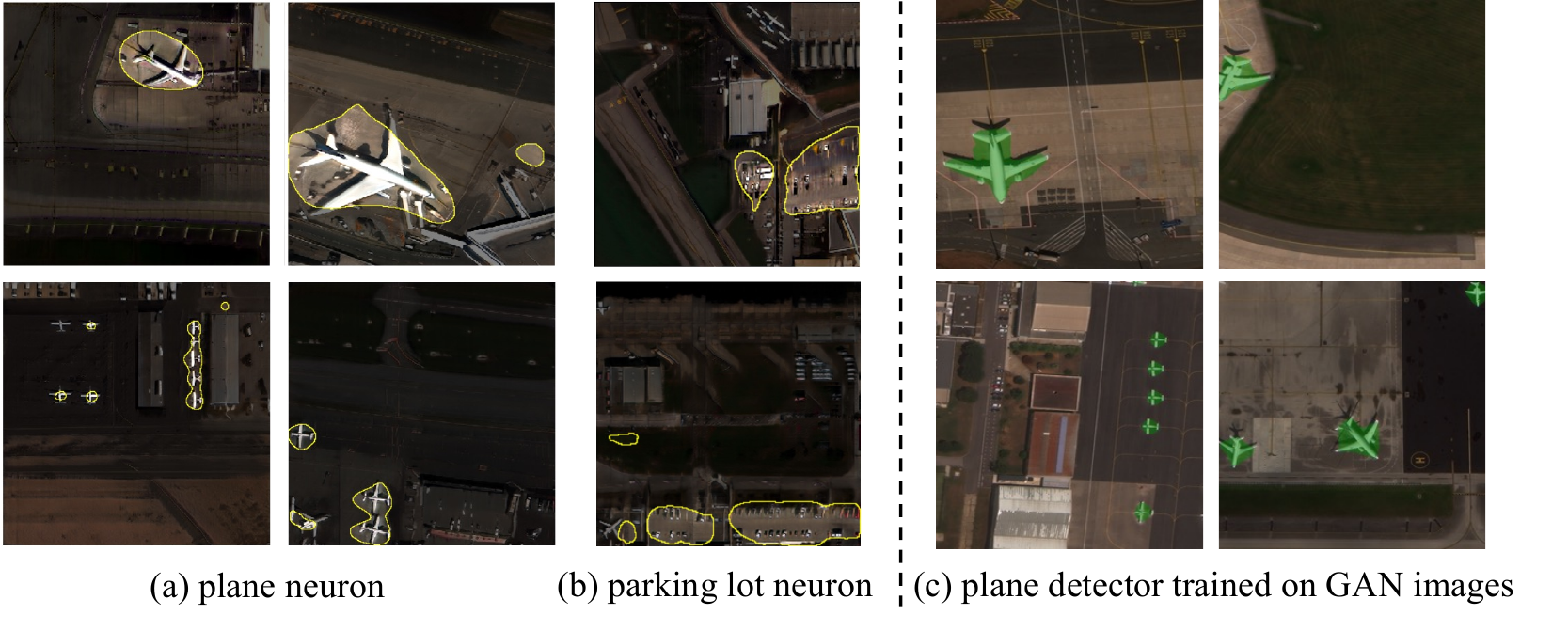}
        \vspace{-5mm}
        \captionof{figure}{
        We leverage the CLIP model to identify neurons in StyleGAN that explicitly express the concept of objects: (a) illustrates the high activation region of the ``plane'' neuron; in addition, (b) illustrates the area highlighted by the ``parking lot'' neuron. The proposed method then distills the knowledge of these neurons and builds object detectors without human annotation. }
        \label{fig:teaser}
    \end{center}
    }]
}

\begin{document}
%
\maketitle
%



\begin{abstract}
We present a fully automated framework for building object detectors on satellite imagery without requiring any human annotation or intervention. We achieve this by leveraging the combined power of modern generative models (e.g., StyleGAN \cite{Karras2020ada}) and recent advances in multi-modal learning (e.g., CLIP \cite{radford2021learning}). While deep generative models effectively encode the key semantics pertinent to a data distribution, this information is not immediately accessible for downstream tasks, such as object detection. In this work, we exploit CLIP's ability to associate image features with text descriptions to identify neurons in the generator network, which are subsequently used to build detectors on-the-fly.
\end{abstract}
\section{Introduction}
The advances of generative modeling in image understanding and computer vision motivate their application on remote sensing and satellite imagery~\cite{RarePlanes_Dataset}. This application space presents unique challenges, where large amount of unlabeled, high-resolution data is available while being constantly updated. However, obtaining annotations and labels often is often prohibitively expensive, not only due to its demand for domain expertise but also its sensitive nature in various applications.  As a result, the efficacy of modern deep architectures in handling high-resolution images (e.g., convolution networks~\cite{he2016identity}, transformers~\cite{vaswani2017attention}) and in supporting representation learning on unlabeled data have had a transformative impact in these applications. While the inherent challenge in curating fine-grained labels for satellite images still persists~\cite{RarePlanes_Dataset}, the use of pre-trained representations have significantly relaxed the annotation requirement for different predictive modeling tasks~\cite{manas2021seasonal}. Despite these advances, the performance of the resulting models tends to be poor when the labeled data is extremely sparse or unavailable in practice. 

In this paper, we investigate the problem of constructing object detectors ``on-the-fly'' -- i.e., a completely zero-shot setting where we cannot access \emph{any} training annotations. Note, in contrast to existing zero-shot learning (ZSL) problems that attempt to extend a classifier to include additional un-sampled classes, our goal is to build a detector for a specific semantic concept (for e.g. \textit{plane}), without requiring the users to provide any examples or explicit specification of the attributes of interest (e.g., shape, color, size). Designing such a zero-shot approach requires two key ingredients: (i) an accurate description of all attributes (local/global) in satellite images using only unlabeled data; 
(ii) ability to align semantic concepts (text description) with image characteristics in a joint latent space. While the former component offers controllability of specific features even in complex satellite imagery, the latter automatically provides a mapping between image features and human-understandable semantic descriptors. Thankfully, recent advances in deep generative models (e.g., StyleGAN~\cite{karras2019style}) and multimodal embedding techniques (e.g., CLIP~\cite{radford2021learning}) make this goal feasible. 

Our approach begins with the user specifying a semantic concept that needs to be detected (i.e., localize the region), which we transform using a pre-trained text encoder from CLIP~\cite{radford2021learning}. Next, we propose a new approach to automatically identify specific neurons from the pre-trained StyleGAN's generator network that are most aligned with the semantic concept. Note, we are able to measure this alignment by projecting the reconstruction from StyleGAN into the joint latent space using the image encoder from CLIP. 

Remarkably, we found that a na\"ive detector that we design based on the activations of the selected StyleGAN's neurons is able to effectively identify different variants of an object without any user guidance (different locations in the image, varying sizes, varying number of occurrences in a scene). Moreover, we showed that curating a ``pseudo''-labeled dataset using our na\"ive detector and training an explicit object segmentation model from scratch (e.g., U-Net) leads to improved performance. Using standard benchmarks, we demonstrate the effectiveness of our zero-shot detector and present comparisons to an \textit{oracle} supervised training. In summary, this work sheds light into how recent large-scale, pre-trained models can be leveraged in lieu of expensive data annotation.

    


\section{Related Work}
Several existing works have demonstrated StyleGAN's abilities to capture physical attributes (size, shape, color, texture etc.) automatically from unsupervised data~\cite{liu2022sparsity,jahanian2019steerability}, but still require human supervision to explicitly label the discovered attributes. Instead of discovering attributes in the latent space, one can also define concepts by characterizing individual neurons' functionality. To this end, Bau \textit{et al.} proposed a neuron interpretation method that utilizes an image dataset with rich segment-level labels \cite{bau2020understanding, bau2018gan}. In the proposed work, we also adopt a similar neuron-level semantic modeling approach. However, instead of relying on the human annotation of objects in the image as in \cite{bau2020understanding, lu2022interactive}, we leverage the recently CLIP multimodal embeddings~\cite{patashnik2021styleclip, kim2021diffusionclip}. Discovering semantically aligned neurons from a StyleGAN will naturally enable us to produce an arbitrarily large ``pseudo''-labeled dataset. Existing works have also adopted a similar approach, but require at least a few annotated examples. For example, the recently proposed datasetGAN~\cite{zhang21} uses a few human annotations of GAN-generated images to construct an ``annotated'' GAN data factory, which is capable of producing a large number of examples with fine-grained labels.

\begin{figure*}[htbp]
    \centering
    \includegraphics[width=\linewidth]{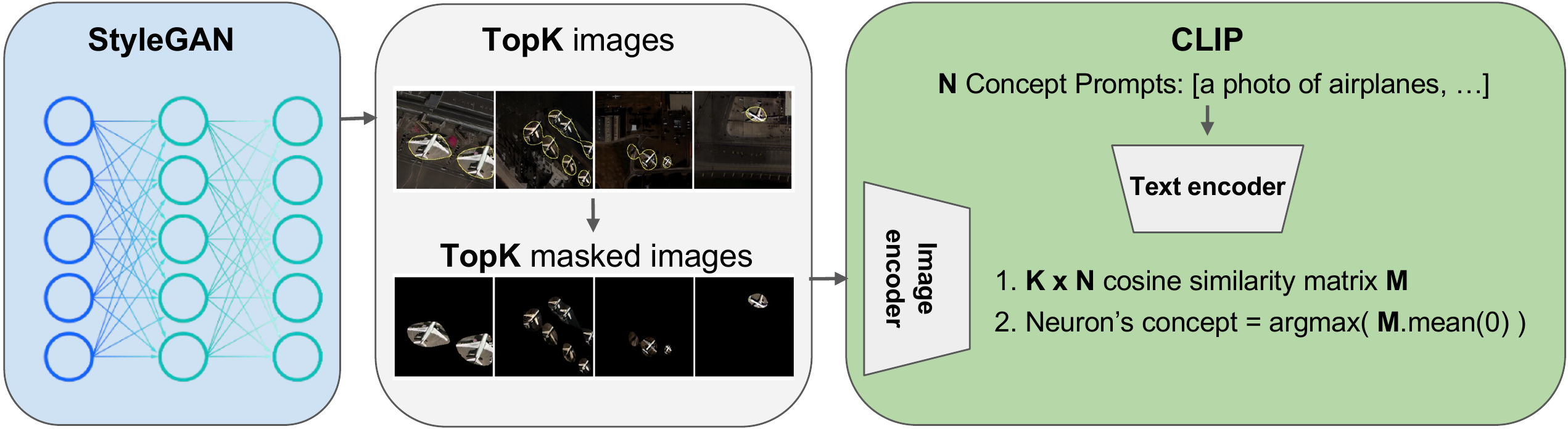}
    \caption{The proposed algorithm for quantifying a neuron's capability as a specific object detector. Given a StyleGAN pretrained on the target domain and a CLIP model with image and text encoders, we first gather topk images for each neuron in all convolutional layers. Then, we mask out regions with low activation values so that only features the neuron focuses on are retained. We feed these masked topk images to CLIP's image encoder and a set of prompts containing objects people are interested in to CLIP's text encoder to get image embeddings $I$ and text embeddings $T$, respectively. Finally, we compute the cosine similarity between $I$ and $T$ to get a similarity matrix $M$. The object a neuron detects is contained in the prompt that has the maximum average similarity with the neuron's topk masked images. }
    \label{fig:method}
\end{figure*}

\section{CLIP-Guided Object-Centric Neuron Discovery in StyleGAN}
\label{sec:method}
One of the key ingredients that facilitates the proposed approach is the expressiveness of generative models (e.g., StyleGAN) and the semantic control they exhibit \cite{bau2020understanding, harkonen2020ganspace, shen2021closed, voynov2020unsupervised}. However, existing unsupervised methods for discovering task-relevant attributes from generative models often produce a large number of unlabeled suggestions, which require human supervision to be leveraged for any downstream task. 
%
In this work, we overcome some of these shortcomings by utilizing the multimodal CLIP model \cite{radford2021learning} to provide guidance. The inherent capability of CLIP to embed both image and language descriptors into a joint feature space enables us to `annotate' each of the StyleGAN neurons (in the generator) by the semantic concepts that they align with. We achieve this in two stages: (i) for each neuron, gather quantile statistics of its activation value and save top activated images and (ii) leverage CLIP's joint embedding space to select neurons that are maximally aligned to a semantic concept of interest.

As illustrated in Fig.\ref{fig:method}, for a given neuron, we first identify the top $\mathrm{K}$ activation images based on maximum activation intensity, $I_k = \mathcal{G}(z_k)$, where $0 < k \leq \mathrm{K}$, from a set of randomly generated images from the GAN $\mathcal{G}$ pretrained on a satellite image dataset. We then isolate and focus on the high-activation regions using a simple mask, $M_k$, following which the masked image $I_k' = M_k \circ I_k$ is embedded into the the CLIP embedding space along with $\mathrm{N}$ candidate prompts (for e.g., for airplanes we use \emph{``an aerial photograph of airplanes''}, i.e., $N = 1$). 

These prompts describe the objects we aim to identify, denoted by $T_n$. The most related concept of the given neuron can then assigned by examining which prompt achieves the largest average cosine similarity between all $\mathrm{K}$ image and text embeddings in the CLIP latent space: $ \sum_{k=1}^\mathrm{K} (Sim(\hat{I}_{k},  \hat{T_{n}})) / \mathrm{K}$, where $\hat{I}_{k} = \mathrm{CLIP}_{image}(I_k')$, $\hat{T}_{n} = \mathrm{CLIP}_{text}(T_n)$, are the corresponding image and text embedding in CLIP model, and $Sim$ is the cosine similarity between the embedding vectors. 


Given a prompt, we can identify the most relevant neuron at each convolutional layer by ranking the average cosine similarities. In practice, we choose the neuron with the highest cosine similarity from each convolutional layer with a feature map size of $16 \times 16$ and above. This way, we leverage neurons with multiple scales. We did not consider neurons in earlier convolutional layers as their feature maps were too coarse. As illustrated in Fig.\ref{fig:teaser}, using the prompts \emph{``an aerial photograph of airplanes''} and \emph{``an aerial photograph of a parking lot''}, our method is able to identify neurons that can localize aircrafts and parking lots respectively.

\begin{table*}[t]
    \caption{The performance of trained U-Nets using different datasets measured by F1 score. The numbers in parentheses are recall and precision, respectively. Here the last row represents the Oracle, i.e., real satellite images with detailed human annotation.}
      \vspace{1ex}
      \centering
        \begin{tabular}{c|c|c c c}
        \hline
        Image & Annotation & Random  & Pretrained \\
        \midrule
        GAN generated images & Neurons & 0.31 (0.39 / 0.26) & 0.30 (0.42 / 0.24) \\
        
        GAN generated images & \small{Neurons + Label Smoothing} & 0.35 (0.43 / 0.27) & 0.30 (0.36 / 0.26) \\
        
        GAN generated images & \small{Neurons + CLIP Consistency} & 0.42 (0.49 / 0.36) & 0.26 (0.29 / 0.23) \\
        \midrule
        Satellite images (\emph{Rareplane} dataset) & Human  & 0.78 (0.81 / 0.74) & 0.82 (0.87 / 0.79) \\
        \hline
        \end{tabular}
      \vspace{1ex}
    \label{tab:results}
\end{table*}


    

\section{Object detection with annotated StyleGAN}
\label{sec:inference}
Once we have identified neurons as relevant object detectors, we can use them to label images. However, utilizing these neurons directly on real images are not feasible without a highly accurate GAN inversion model, i.e., during inference time, for an unseen image, we first invert it into the GAN's latent space, and then using the identified the detector neuron (see Section \ref{sec:method}) we can instantaneously identify the object and highlight its location. 
While current state-of-the-art GAN inversion models such as \cite{richardson2021encoding} work extremely well for aligned images such as faces, we found through experimentation that they fail on satellite images, which are unaligned and much more complex. As a result, in order to distill the knowledge of the detector neuron, we trained a separate detector by utilizing the StyleGAN generated images along with the corresponding annotation generated solely by neurons . 

For the detector, we used an U-Net \cite{ronneberger2015u}, a classic architecture consisting of an encoder to extract image features and a symmetric decoder to enable precise localization, as our downstream segmentation model. The encoder we used is ResNet18\cite{he2016identity}. We split the original Rareplanes \cite{RarePlanes_Dataset} training set into 5233 training images and 582 validation images while keeping the test set of size 2710 intact; the training images and corresponding label annotations are never used during training our model. Correspondingly, we sampled 5233 random images from GAN and gather their annotations generated by neurons. 

Since these neuron-generated annotations are inherently noisy, we adopted two methods to improve their quality: (a) \textbf{Label smoothing:} is a widely used method \cite{pereyra2017regularizing} to tackle the overconfidence of neural networks, that adds a small penalty to encourage predictions with higher entropy; and (b) \textbf{Label consistency with CLIP}: we discard images whose masked version based on neuron-annotation have low cosine similarity ($<0.24$) with prompts in CLIP's embedding space and keep sampling until reaching the target training set size. This cosine similarity threshold ($0.24$) is the mean similarity between a masked training set and the prompt for airplanes. We emphasize that generating these datasets and corresponding annotations requires no form of human supervision and is completely automated. We used dice loss \cite{milletari2016v} for training the U-Net model, and models with the best validation loss were selected for testing. 

During testing, we convert both the predicted and ground truth segmentations to the tightest bounding box and evaluate models' performance by F1 score, setting prediction threshold to IoU $>$ $0.5$. We opted in this evaluation protocol because for satellite images, localization is often sufficient. To establish a performance reference, we trained a fully supervised model using 5233 training samples from Rareplanes dataset. In all experiments, we trained the model using Adam optimizer and a learning rate of 0.001 for 20 epochs.

\noindent \textbf{Results:} Our experiment results are summarized in table \ref{tab:results}. The first column indicates the source of the training images. The Oracle utilizes the original real images, whereas the experiments of the proposed method trained on GAN generate images don't. The second column shows what supervision is provided for training. For the Oracle, detailed human annotations are provided. For the proposed method, we experimented with several variations of neuron annotation discussed previously: 1) use only neuron annotation in GAN; 2) augment neuron annotation with label smoothing; 3) augment neuron annotation with label consistency with CLIP.
We can see that the proposed method achieved outstanding performance under the restrictive problem setting, i.e., utilizing only GAN-generated images and no human annotations or input. The performance gap between the Oracle and the proposed method is expected, as the Oracle represents the most optimal and more importantly supervised setting. 

\begin{figure}[htbp]
    \centering
    \includegraphics[width=0.8\linewidth]{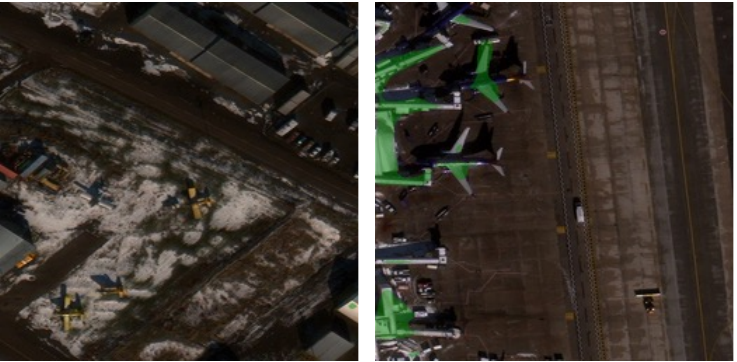}
    \caption{Here we show failure cases of our trained U-Net. In left example, the model missed airplanes that are blended in the background. In the right example, the model mistakes part of terminals for airplanes.}
    \label{fig:failure_cases}
\end{figure}

When training U-Net with random initialization, both label smoothing and label consistency alleviate the noisiness in the training set and help improve performance, with label consistency by CLIP boosting the F1 score by 0.11. However, when starting with pre-trained weights, the proposed methods suffer from a performance drop. We hypothesize that there is a gap between GAN-generated images and real images, and label consistency by CLIP seems to further amplify the difference. This problem can be alleviated if we improve the quality of the GAN by including more training datasets and/or increasing its resolution. 
Lastly, we also include examples of failure cases. As we can see in Fig. \ref{fig:failure_cases}), the complex and snowy background hides the planes in the left image. And the model mistaken part of the terminals, which is also white, as airplanes.

\section{Discussion and Future Work}
In this work, we propose a novel approach for building object detectors i in a challenging application domain without utilizing any human annotation or supervision. To the best of our knowledge, this work demonstrates the first effort to directly annotate neurons with the help of another neural network model (CLIP). By removing human involvement in the predictive pipeline, we achieved an on-the-fly zero-shot setup that can accommodate unseen objects in an unknown dataset. However, as we rely on CLIP for identifying objects of interest, we also inherit the potential limitation and bias of the CLIP model. The CLIP model is trained on a large number of natural images that are potentially different from the top-down satellite images, so not all prior knowledge is transferable. To mitigate this issue, we adopted a CLIP model that is fine-tuned on satellite images (the original CLIP model produces slightly worse but still comparable results). 
For future directions, we plan to explore the GAN-based knowledge mining approach and experiment with the proposed methods on more diverse datasets and other multimodal models. Even though in this work we only focused on satellite imagery, the proposed method can be adopted for other applications and dataset and potentially work better as satellite images present many unique challenges that are not seen elsewhere.
Moreover, we also plan to perform more thorough analysis on failure cases to understand the performance drop caused by using pretrained weights. 

\section{Acknowledgement}
This work performed under the auspices of the U.S. Department of Energy by Lawrence Livermore National Laboratory under Contract DE-AC52-07NA27344. 

%



\bibliographystyle{IEEEbib}
\bibliography{refs}

\begin{thebibliography}{10}

\bibitem{Karras2020ada}
Tero Karras, Miika Aittala, Janne Hellsten, Samuli Laine, Jaakko Lehtinen, and
  Timo Aila,
\newblock ``Training generative adversarial networks with limited data,''
\newblock in {\em Proc. NeurIPS}, 2020.

\bibitem{radford2021learning}
Alec Radford, Jong~Wook Kim, Chris Hallacy, Aditya Ramesh, Gabriel Goh,
  Sandhini Agarwal, Girish Sastry, Amanda Askell, Pamela Mishkin, Jack Clark,
  et~al.,
\newblock ``Learning transferable visual models from natural language
  supervision,''
\newblock in {\em International Conference on Machine Learning}. PMLR, 2021,
  pp. 8748--8763.

\bibitem{RarePlanes_Dataset}
Jacob Shermeyer, Thomas Hossler, Adam Van~Etten, Daniel Hogan, Ryan Lewis, and
  Daeil Kim,
\newblock ``Rareplanes dataset,'' June 2020.

\bibitem{he2016identity}
Kaiming He, Xiangyu Zhang, Shaoqing Ren, and Jian Sun,
\newblock ``Identity mappings in deep residual networks,''
\newblock in {\em European conference on computer vision}. Springer, 2016, pp.
  630--645.

\bibitem{vaswani2017attention}
Ashish Vaswani, Noam Shazeer, Niki Parmar, Jakob Uszkoreit, Llion Jones,
  Aidan~N Gomez, {\L}ukasz Kaiser, and Illia Polosukhin,
\newblock ``Attention is all you need,''
\newblock {\em Advances in neural information processing systems}, vol. 30,
  2017.

\bibitem{manas2021seasonal}
Oscar Manas, Alexandre Lacoste, Xavier Gir{\'o}-i Nieto, David Vazquez, and Pau
  Rodriguez,
\newblock ``Seasonal contrast: Unsupervised pre-training from uncurated remote
  sensing data,''
\newblock in {\em Proceedings of the IEEE/CVF International Conference on
  Computer Vision}, 2021, pp. 9414--9423.

\bibitem{karras2019style}
Tero Karras, Samuli Laine, and Timo Aila,
\newblock ``A style-based generator architecture for generative adversarial
  networks,''
\newblock in {\em Proceedings of the IEEE/CVF Conference on Computer Vision and
  Pattern Recognition}, 2019, pp. 4401--4410.

\bibitem{liu2022sparsity}
Shusen Liu, Rushil Anirudh, Jayaraman~J Thiagarajan, and Peer-Timo Bremer,
\newblock ``Sparsity improves unsupervised attribute discovery in stylegan,''
\newblock in {\em ICASSP 2022-2022 IEEE International Conference on Acoustics,
  Speech and Signal Processing (ICASSP)}. IEEE, 2022, pp. 3388--3392.

\bibitem{jahanian2019steerability}
Ali Jahanian, Lucy Chai, and Phillip Isola,
\newblock ``On the "steerability" of generative adversarial networks,''
\newblock {\em arXiv preprint arXiv:1907.07171}, 2019.

\bibitem{bau2020understanding}
David Bau, Jun-Yan Zhu, Hendrik Strobelt, Agata Lapedriza, Bolei Zhou, and
  Antonio Torralba,
\newblock ``Understanding the role of individual units in a deep neural
  network,''
\newblock {\em Proceedings of the National Academy of Sciences}, vol. 117, no.
  48, pp. 30071--30078, 2020.

\bibitem{bau2018gan}
David Bau, Jun-Yan Zhu, Hendrik Strobelt, Bolei Zhou, Joshua~B. Tenenbaum,
  William~T. Freeman, and Antonio Torralba,
\newblock ``Visualizing and understanding generative adversarial networks,''
\newblock in {\em International Conference on Learning Representations}, 2019.

\bibitem{lu2022interactive}
Yuzhe Lu and Adam Perer,
\newblock ``An interactive interpretability system for breast cancer screening
  with deep learning,''
\newblock {\em arXiv preprint arXiv:2210.08979}, 2022.

\bibitem{patashnik2021styleclip}
Or~Patashnik, Zongze Wu, Eli Shechtman, Daniel Cohen-Or, and Dani Lischinski,
\newblock ``Styleclip: Text-driven manipulation of stylegan imagery,''
\newblock in {\em Proceedings of the IEEE/CVF International Conference on
  Computer Vision}, 2021, pp. 2085--2094.

\bibitem{kim2021diffusionclip}
Gwanghyun Kim and Jong~Chul Ye,
\newblock ``Diffusionclip: Text-guided image manipulation using diffusion
  models,''
\newblock 2021.

\bibitem{zhang21}
Yuxuan Zhang, Huan Ling, Jun Gao, Kangxue Yin, Jean-Francois Lafleche, Adela
  Barriuso, Antonio Torralba, and Sanja Fidler,
\newblock ``Datasetgan: Efficient labeled data factory with minimal human
  effort,''
\newblock in {\em CVPR}, 2021.

\bibitem{harkonen2020ganspace}
Erik H{\"a}rk{\"o}nen, Aaron Hertzmann, Jaakko Lehtinen, and Sylvain Paris,
\newblock ``G{AN}space: Discovering interpretable {GAN} controls,''
\newblock {\em arXiv preprint arXiv:2004.02546}, 2020.

\bibitem{shen2021closed}
Yujun Shen and Bolei Zhou,
\newblock ``Closed-form factorization of latent semantics in {GAN}s,''
\newblock in {\em Proceedings of the IEEE/CVF Conference on Computer Vision and
  Pattern Recognition}, 2021, pp. 1532--1540.

\bibitem{voynov2020unsupervised}
Andrey Voynov and Artem Babenko,
\newblock ``Unsupervised discovery of interpretable directions in the {GAN}
  latent space,''
\newblock in {\em International Conference on Machine Learning}. PMLR, 2020,
  pp. 9786--9796.

\bibitem{richardson2021encoding}
Elad Richardson, Yuval Alaluf, Or~Patashnik, Yotam Nitzan, Yaniv Azar, Stav
  Shapiro, and Daniel Cohen-Or,
\newblock ``Encoding in style: a stylegan encoder for image-to-image
  translation,''
\newblock in {\em Proceedings of the IEEE/CVF conference on computer vision and
  pattern recognition}, 2021, pp. 2287--2296.

\bibitem{ronneberger2015u}
Olaf Ronneberger, Philipp Fischer, and Thomas Brox,
\newblock ``U-net: Convolutional networks for biomedical image segmentation,''
\newblock in {\em International Conference on Medical image computing and
  computer-assisted intervention}. Springer, 2015, pp. 234--241.

\bibitem{pereyra2017regularizing}
Gabriel Pereyra, George Tucker, Jan Chorowski, {\L}ukasz Kaiser, and Geoffrey
  Hinton,
\newblock ``Regularizing neural networks by penalizing confident output
  distributions,''
\newblock {\em arXiv preprint arXiv:1701.06548}, 2017.

\bibitem{milletari2016v}
Fausto Milletari, Nassir Navab, and Seyed-Ahmad Ahmadi,
\newblock ``V-net: Fully convolutional neural networks for volumetric medical
  image segmentation,''
\newblock in {\em 2016 fourth international conference on 3D vision (3DV)}.
  IEEE, 2016, pp. 565--571.

\end{thebibliography}

\end{document}